# Categorising Products in an Online Marketplace: An Ensemble Approach

Kieron Drumm, *School of Computing, Dublin City University, Dublin, Ireland*

In recent years, product categorisation has been a common issue for E-commerce companies who have utilised machine learning to categorise their products automatically. In this study, we propose an ensemble approach, using a combination of different models to separately predict each product's category, subcategory, and colour before ultimately combining the resultant predictions for each product. With the aforementioned approach, we show that an average F1-score of 0.82 can be achieved using a combination of XGBoost and k-nearest neighbours to predict said features.

## 1. INTRODUCTION

Modern E-commerce platforms, such as Amazon or Rakuten, store information about millions of different products, all available on their online marketplaces in digital catalogues and commonly represented with multi-level taxonomy trees with thousands of leaf nodes [7]. The maintenance of these tree structures can be tedious, depending on the variety of products offered or the rate at which new products are added to the catalogues of these E-commerce platforms. Furthermore, the manual categorisation of new products can often be inaccurate and inconsistent, as it can depend on the merchant to whom the product listings belong. To rectify these shortcomings, automatic categorisation has become popular in recent years, as it provides a more feasible means of keeping an extensive catalogue up-to-date [7]. The following work outlines a method for predicting top-level and sub-level categories and product colour on a popular E-commerce platform when provided additional information about said products, such as the title, description, or aesthetic details, such as the object's style, shape, or pattern.

## 2. RELATED WORK

### 2.1 Imbalanced Data

A common issue with most modern machine learning algorithms is that they assume that the data they are being trained on is balanced [1]. This assumption can be detrimental to the performance of the models trained with these algorithms, as imbalances in the data can result in the predictions of said models being biased and skewed toward the majority class [1].

Some have suggested that the optimal means of resolving imbalances lies with oversampling techniques, such as the Synthetic Minority Oversampling Technique (SMOTE) or the arguably improved form of SMOTE: Adaptive Synthetic (ADASYN) [1]. Research into the use of such techniques has shown that when comparing the effectiveness of machine learning models trained using algorithms such as logistic regression, support vector machine, Random Forest, and XGBoost, the application of either SMOTE or ADASYN results in a significant increase in the F1-score and AUC-ROC [1]. For example, in a study carried out by [1], a Random Forest model trained on an untouched dataset was shown to produce an F1-score of 0.18 and an AUC-ROC of 0.55, whereas a model trained on a newly balanced dataset was shown to produce an F1-score of 0.91 and an AUC-ROC of 0.91 [13]. However, it should be noted that [1] also acknowledged the shortcomings of SMOTE, namely the fact that it tends to intensify noise in the data and its tendency to oversample minority instances with a uniform likelihood [1].

Others have suggested the use of ensemble algorithms as a means of dealing with imbalances. One such approach was posed by [2], in which they presented the use of undersampling, cost-sensitive learning, bagging, and support vector machine (SVM) to train an effective model on imbalanced medical data. Through comparative analysis, they were able to show that this novel approach could outperform commonly used algorithms, such as ADABoost, Random Forest, EasyEnsemble, logistic regression, and k-nearest neighbours [2], whereas [3] proposed a combination of Borderline-SMOTE and an ensemble of SVM models, aptly named: Bagging of Extrapolation Borderline-SMOTE SVM (BEBS), as they recognised the importance of placing value on samples in the data that exist near decision boundaries [3].

2.2 Implementation

Research into approaches for large-scale product categorisation has become increasingly popular in recent years. Product categorisation has been, for the most part, treated as a standard classification task in which information about a given product is provided as input, and the output is the most likely category into which the product falls [4]. This form of classification has also been approached in two distinct ways: a single-step approach, in which a product is classified into one of a large number of categories in a "single step", or the step-wise approach, in which a product's top-level category is first predicted, after which said prediction is used to predict that same product's subcategory [4].

However, some researchers have also approached product categorisation in novel ways. One such approach can be seen in the work of [4], who chose to treat the aforementioned problem as a natural language processing (NLP) task and subsequently made use of a neural machine translation (NMT) model to translate a product's natural language description into a sequence of tokens representing a root-to-leaf path in a product taxonomy [4]. The NMT model developed made use of previously trained models, such as the attentional Seq2Seq model, proposed by [5], and the Transformer model, proposed by [6], and was able to achieve a weighted F1-score of 74.94 when trained on a dataset with a 60-10-30 (train-validation-test) split; a result that outperformed previous attempts at performing product categorisation with the use of NLP.

Another novel approach proposed in recent years can be seen in the work of [7], who proposed the use of an attention convolution neural network (ACNN) to impute attentive focus on specific word tokens in the titles of products. This model was inspired by an attention-based CNN model previously proposed by [8]. With regards to results, not only was this approach able to achieve an improved macro F1-score than that of a gradient-boosted tree (GBT), but, interestingly, this attention-based approach also showed that the words chosen by the attention mechanism were semantically correlated with the category of the product to which they were linked [7].

## 3. METHODOLOGY

### 3.1 Data Transformation

#### 3.1.1 Data Imbalances

It has been observed that any data that is organised into higher and lower-level categories is subject to what is commonly referred to as the "long tail phenomenon" [4]: a data imbalance that occurs when a large proportion of products in a catalogue are distributed over a small number of categories, leaving the remaining fraction of products to be sprinkled over a large number of remaining categories [4]. The dataset used in this study was no exception, as clear imbalances in both the high and lower-level categories were identified during the initial analysis phase. To resolve this imbalance in the data, a combination of oversampling, using the Synthetic Minority Oversampling Technique (SMOTE) [1], and random undersampling was utilised. This combination ensured that bias was effectively reduced in the majority classes in the dataset while also ensuring that bias was improved in the minority classes.

#### 3.1.2 Features with High Cardinality

When dealing with categorical features with a high level of cardinality, it can be challenging to encode said features effectively, as many standard encoding techniques, such as one-hot encoding, can suffer due to their tendency to create a large number of output features or their inability to capture morphological information when dealing specifically with strings [9]. To encode the string features encountered during this study, the Min-Hash encoding technique was used, as it has been shown to perform well on features with high cardinality due to its ability to produce low dimensional representations of said features while still being scalable [9].

#### 3.1.3 Missing Values

Missing values in datasets often occur for one of three reasons: missing completely at random (MCAR): when the probability of missing data on a variable is unrelated to any other measured variable and is unrelated to the variable with missing values itself; missing at random (MAR): when the probability of missing data on a variable is related to some other measured variable in the model; or not missing at random (NMAR): when the missing values on a variable are related to the values of that variable itself [10]. Several features in this dataset contained many missing values, some reaching over 90% in terms of the amount missing. Due to the sheer scale of missing data in the dataset, the possibility of simply dropping any affected columns was considered but subsequently discarded, as removing the majority of the features in the dataset would have resulted in far too much information being lost. With this in mind, the fast-KNN (fast-K-nearest neighbours) algorithm, an optimised form of KNN imputation, was employed for the purposes of imputation [18]. This algorithm was chosen because it has been shown to perform better relative to the standard sample and median imputation methods [18].

3.2 Models

3.2.1 K-nearest Neighbours

K-nearest neighbours (KNN) is a non-parametric, distance-based, supervised learning model commonly used for regression and classification tasks. It makes predictions by regarding the distances between new inputs and known training samples as a measure of similarity [11]. A new input is classified by first calculating the distance between the input and all known training samples, using either Euclidean distance, Manhattan distance, Minkowski distance, or a number of other popular distance metrics [11]. Once all distances have been calculated, the cluster deemed as being closest to the new point is chosen, and the class to which that cluster belongs is output as the predicted class for the input [12]. Though it can suffer when run against datasets with a high dimensionality [13], it was chosen due to its ability to handle features with large numbers of classes. When training a model to predict the "bottom_category" and "color" features, an "n_neighbors" value of 1 and the "Manhattan" distance metric were used as they were observed to produce the optimal result on a sample size of 25,000 (see Table 1).

**Table 1. Hyperparameters for the K-nearest Neighbours Model.**

| Parameter Name | Value |
| --- | --- |
| metric | "manhattan" |
| n_neighbors | 1 |

3.2.2 Random Forest

Random Forest is a tree-based, supervised learning model composed of many individual decision trees operating as an ensemble [14]. It performs classification by summing up the outputs from each of the individual trees in the "forest" and subsequently determines the class with the highest number of votes, which is then output as the final classification for the model [14]. For this study, a Random Forest classifier was trained on a sample size of 10,000 with 50 separate trees, a max tree depth of 9, and all impurity was measured using Gini impurity due to its greedy nature and ability to train models in a short amount of time [15]. All other hyperparameters may be seen in Table 2.

**Table 2. Hyperparameters for the Random Forest Model.**

| Parameter Name | Value |
| --- | --- |
| max_depth | 9 |
| max_features | "log2" |
| n_estimators | 50 |
| n_jobs | 1 |
| oob_score | True |
| random_state | 411 |

3.2.3 XGBoost

XGBoost is a tree-based, supervised learning model that has been shown to perform better than other popular models when it comes to supervised learning on structured data [16]. Though it may be similar to Random Forest in that it uses decision trees, a key difference is that it utilises gradient descent and boosting to train all decision trees sequentially, as opposed to the bagging approach utilised by Random Forest that enables training in parallel [17]. When training a model to predict the "top_category" feature, a learning rate of 0.3 and a min_split_loss of 0.1 were chosen, along with several other optimal parameters (see Table 3).

Table 3. Hyperparameters for the XGBoost Model.

| Parameter Name | Value |
|---|---|
| eval_metric | "error" |
| learning_rate | 0.3 |
| min_split_loss | 0.1 |
| objective | "multi:softprob" |
| predictor | "auto" |
| tree_method | "auto" |

## 4. EXPERIMENTS & RESULTS

### 4.1 Dataset

The data used in this study came in the form of a collection of either Parquet or TFRecord files and was supplied by a popular E-commerce company whose name shall remain anonymous due to confidentiality reasons. The Parquet form of the dataset contained a great deal of information about a number of products sold by this company, including the occasion for which it was meant, the top and bottom level categories of the product, and a title and description for the product, along with several other features. In contrast, the TFRecord form of the dataset contained images of each product and other features describing the product. For this study, the Parquet dataset was chosen.

### 4.2 Evaluation Metrics

To evaluate each of the aforementioned classifiers, the precision, recall, and F1-score metrics were used. The precision and recall were calculated as an initial measure of how well each classifier had performed, after which the F1-Score for each result was calculated due to its ability to elegantly summarise the predictive performance of a model using both precision and recall [19]. These three metrics were calculated using the following formulae:

$$Precision = \frac{TP}{TP + FP}$$

$$Recall = \frac{TP}{TP + FN}$$

$$F1\text{-}Score = \frac{2 \times Precision \times Recall}{Precision + Recall}$$

Where TP is the number of true positives in the prediction results, FP is the number of false positives, and FN is the number of false negatives.

### 4.3 Results

For each of the three models used in this study: k-nearest neighbours, Random Forest, and XGboost, a number of models were trained on varying sample sizes to ascertain the optimal model for predicting each of the three target features: top_category, bottom_category, and color. When predicting the top_category feature, the XGBoost model was seen to perform best, achieving a precision and recall of 91% and an F1-score of 0.91, followed closely in second by k-nearest neighbours, which achieved a recall of 90%, a precision of 89%, and an F1-score of 0.89. Though the Random Forest model achieved similar results, it was ultimately beaten out by both models (see Figure 1).

However, when attempting to predict the bottom_category feature, the XGBoost and Random Forest models dwindled, achieving a precision and recall of 2% and an F1-score of 0.02 in the case of XBoost, and a precision of 21%, a recall of 23%, and an F1-score of 0.18, in the case of Random Forest; whereas, the k-nearest neighbours model appeared to triumph, predicting the bottom_category with a precision of 81%, a recall of 80%, and an F1-score of 0.77 (see Figure 2).

Lastly, when predicting the color feature, all three models performed well, with the k-nearest neighbours ultimately triumphing over the other two, achieving a precision of 76%, a recall of 79%, and an F1-score of 0.77 (see Figure 3). With the above results in mind, an ensemble was constructed to predict the three features mentioned above on any new samples, with the k-nearest neighbours model predicting both the bottom_category and color features and the XGBoost model predicting the top_cateogory feature.

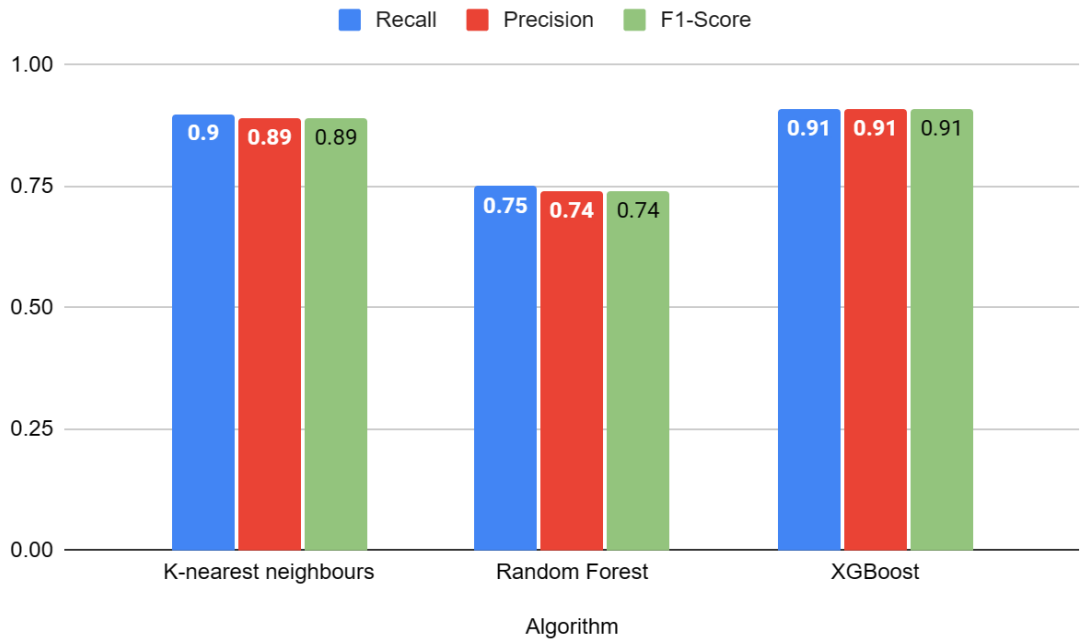

**Figure 1. A Comparison of Prediction Results for the "Top Category" Feature.**

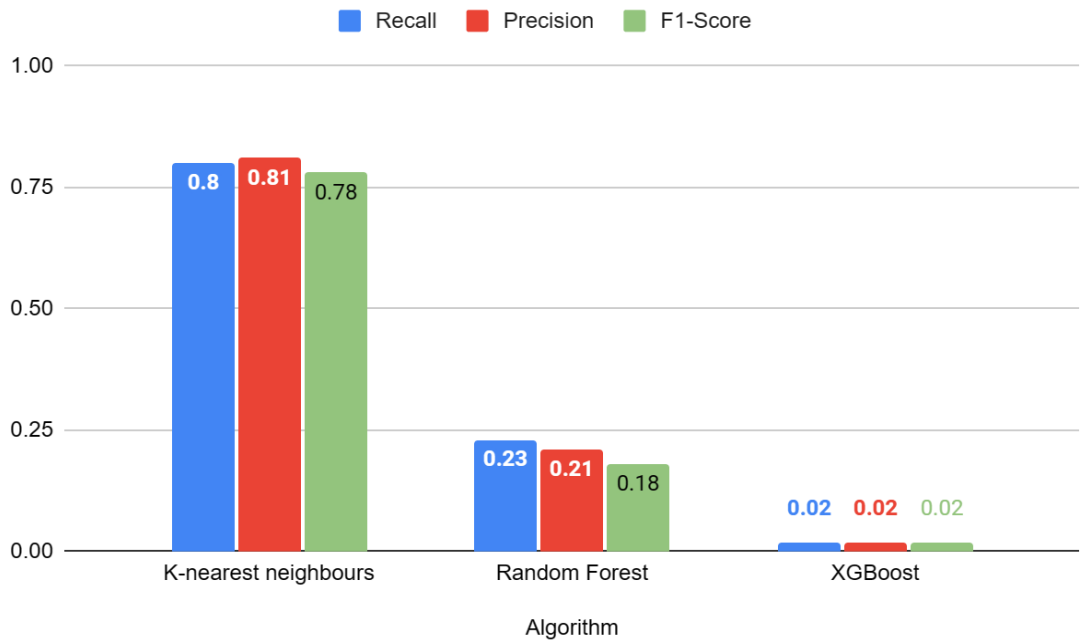

**Figure 2. A Comparison of Prediction Results for the "Bottom Category" Feature.**

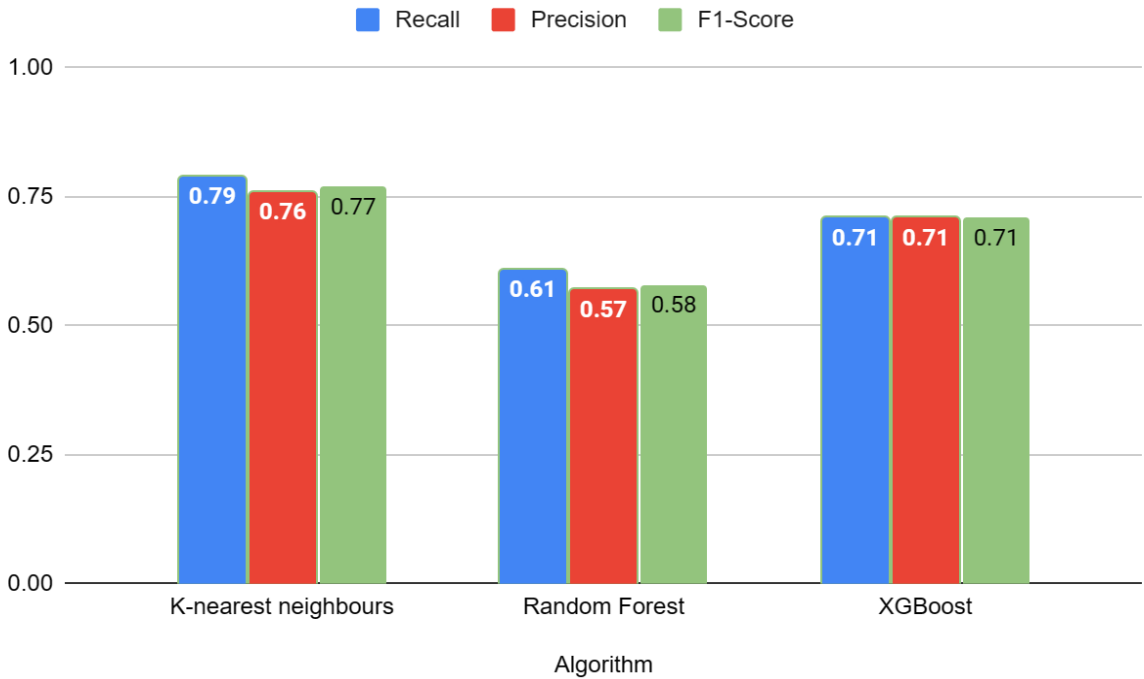

**Figure 3. A Comparison of Prediction Results for the "Color" Feature.**

4.4 Discussion

Our study has shown that an ensemble approach, utilising the intuitive concept of binary relevance [20], achieves the best overall results when dealing with an imbalanced dataset and attempting to predict several hierarchical categorical features. Though satisfactory for the task at hand, we believe that the results achieved could still be improved using more advanced multi-label classification techniques. For example, in order to better utilise the parent-child relationship between the top_category and bottom_category features, a more state-of-the-art approach, such as classifier chains, could be employed, as this technique's ability to chain together binary classifiers to take advantage of label correlations could be highly advantageous [21, 22]. Alternatively, though it is computationally expensive, the Label Powerset approach could also be taken, as it too can take label correlations into account [22].

5. CONCLUSION

In this study, we proposed an ensemble approach for performing multi-label classification on products in an E-commerce catalogue. For each proposed model, we calculated the precision, recall, and F1-score on varying sample sizes: 10,000 for Random Forest and XGBoost, and 25,000 for k-nearest neighbours. When predicting the top_category, bottom_category, and color features, F1-scores of 0.91, 0.78, and 0.77 were achieved, respectively, using a combination of the two best-performing models: XGBoost and k-nearest neighbours. As the test dataset used in this study did not contain the ground truth for product categories and colours, the results reported here all pertain to the training dataset.

REFERENCES

[1] R. Addo Danquah, Handling Imbalanced Data: A Case Study for Binary Class Problems. 2020. doi: 10.6084/m9.figshare.13082573.
[2] L. Liu, X. Wu, S. Li, Y. Li, S. Tan, and Y. Bai, 'Solving the class imbalance problem using ensemble algorithm: application of screening for aortic dissection', BMC Medical Informatics and Decision Making, vol. 22, no. 1, p. 82, Mar. 2022, doi: 10.1186/s12911-022-01821-w.
[3] Q. Wang, Z. Luo, J. Huang, Y. Feng, and Z. Liu, 'A Novel Ensemble Method for Imbalanced Data Learning: Bagging of Extrapolation-SMOTE SVM', Computational Intelligence and Neuroscience, vol. 2017, p. e1827016, Jan. 2017, doi: 10.1155/2017/1827016.
[4] M. Y. Li, S. Kok, and L. Tan, 'Don't Classify, Translate: Multi-Level E-Commerce Product Categorization Via Machine Translation'. arXiv, Dec. 13, 2018. Accessed: Apr. 11, 2023. [Online]. Available: http://arxiv.org/abs/1812.05774.

[5] M.-T. Luong, H. Pham, and C. D. Manning, 'Effective Approaches to Attention-based Neural Machine Translation'. arXiv, Sep. 20, 2015. doi: 10.48550/arXiv.1508.04025.

[6] A. Vaswani et al., 'Attention Is All You Need'. arXiv, Dec. 05, 2017. doi: 10.48550/arXiv.1706.03762.

[7] Y. Xia, A. Levine, P. Das, G. Di Fabbrizio, K. Shinzato, and A. Datta, 'Large-Scale Categorization of Japanese Product Titles Using Neural Attention Models', in Proceedings of the 15th Conference of the European Chapter of the Association for Computational Linguistics: Volume 2, Short Papers, Valencia, Spain: Association for Computational Linguistics, Apr. 2017, pp. 663–668. Accessed: Apr. 11, 2023. [Online]. Available: https://aclanthology.org/E17-2105.

[8] Z. Yang, D. Yang, C. Dyer, X. He, A. Smola, and E. Hovy, 'Hierarchical Attention Networks for Document Classification', in Proceedings of the 2016 Conference of the North American Chapter of the Association for Computational Linguistics: Human Language Technologies, San Diego, California: Association for Computational Linguistics, 2016, pp. 1480–1489. doi: 10.18653/v1/N16-1174.

[9] P. Cerda and G. Varoquaux, 'Encoding High-Cardinality String Categorical Variables', IEEE Trans. Knowl. Data Eng., vol. 34, no. 3, pp. 1164–1176, Mar. 2022, doi: 10.1109/TKDE.2020.2992529.

[10] I. E. PhD, 'Missing data mechanisms', Iris Eekhout, Jun. 28, 2022. https://www.iriseekhout.com/post/2022-06-28-missingdatamechanisms/ (accessed Apr. 13, 2023).

[11] 'K-Nearest Neighbor (KNN) Explained', Pinecone. https://www.pinecone.io/learn/k-nearest-neighbor/ (accessed Apr. 14, 2023).

[12] A. Christopher, 'K-Nearest Neighbor', The Startup, Feb. 03, 2021. https://medium.com/swlh/k-nearest-neighbor-ca2593d7a3c4 (accessed Apr. 14, 2023).

[13] A. A. Tokuç, 'k-Nearest Neighbors and High Dimensional Data | Baeldung on Computer Science', Jul. 31, 2021. https://www.baeldung.com/cs/k-nearest-neighbors (accessed Mar. 29, 2023).

[14] T. Yiu, 'Understanding Random Forest', Medium, Sep. 29, 2021. https://towardsdatascience.com/understanding-random-forest-58381e0602d2 (accessed Mar. 06, 2023).

[15] J. Lawless, 'Under the Hood: Using Gini impurity to your advantage in Decision Tree Classifiers', Medium, Dec. 30, 2020. https://towardsdatascience.com/under-the-hood-using-gini-impurity-to-your-advantage-in-decision-tree-classifiers-9be030a650d5 (accessed Mar. 29, 2023).

[16] V. Morde, 'XGBoost Algorithm: Long May She Reign!', Medium, Apr. 08, 2019. https://towardsdatascience.com/https-medium-com-vishalmorde-xgboost-algorithm-long-she-may-rein-edd9f99be63d (accessed Apr. 11, 2023).

[17] A. Gupta, 'XGBoost versus Random Forest', Geek Culture, Jun. 01, 2021. https://medium.com/geekculture/xgboost-versus-random-forest-898e42870f30 (accessed Apr. 11, 2023).

[18] M. Mohammed et al., 'Comparison of five imputation methods in handling missing data in a continuous frequency table', AIP Conference Proceedings, vol. 040009, pp. 0400061–0400069, May 2021, doi: 10.1063/5.0053286.

[19] Z. LT, 'Essential Things You Need to Know About F1-Score', Medium, Feb. 25, 2022. https://towardsdatascience.com/essential-things-you-need-to-know-about-f1-score-dbd973bf1a3 (accessed Apr. 14, 2023).

[20] M.-L. Zhang, Y.-K. Li, X.-Y. Liu, and X. Geng, 'Binary relevance for multi-label learning: an overview', Front. Comput. Sci., vol. 12, no. 2, pp. 191–202, Apr. 2018, doi: 10.1007/s11704-017-7031-7.

[21] J. Read, B. Pfahringer, G. Holmes, and E. Frank, 'Classifier Chains: A Review and Perspectives', jair, vol. 70, pp. 683–718, Feb. 2021, doi: 10.1613/jair.1.12376.

[22] K. Nooney, 'Deep dive into multi-label classification..! (With detailed Case Study)', Medium, Feb. 12, 2019. https://towardsdatascience.com/journey-to-the-center-of-multi-label-classification-384c40229bff (accessed Apr. 14, 2023).